\documentclass[11pt, a4paper, logo, copyright]{googledeepmind}

\usepackage[authoryear, sort&compress, round]{natbib}
\usepackage{graphicx} 
\usepackage{subcaption} 
\usepackage{fancyhdr} 
\usepackage{lipsum} 
\usepackage{caption} 
\usepackage{color}

\usepackage{cleveref}
\usepackage{afterpage}

\usepackage{booktabs}

\usepackage{threeparttable}

\title{A High-Throughput Platform to Bench Test Smartphone-Based Heart Rate Measurements Derived From Video}

\correspondingauthor{$mingzher$@google.com}

\author[$\dagger$,1]{Ming-Zher Poh}
\author[1]{Jonathan Wang}
\author[1]{Jonathan Hsu}
\author[1]{Lawrence Cai}
\author[1]{Eric Teasley}
\author[1]{James A. Taylor}
\author[1]{Jameson K. Rogers}
\author[1]{Anupam Pathak}
\author[1]{Shwetak Patel}

\affil[$\dagger$]{Corresponding Author}
\affil[1]{Google Research}

\begin{abstract}
Smartphone-based heart rate (HR) monitoring apps using finger-over-camera photoplethysmography (PPG) face significant challenges in performance evaluation and device compatibility due to device variability and fragmentation. Manual testing is impractical, and standardized methods are lacking. This paper presents a novel, high-throughput bench-testing platform to address this critical need. We designed a system comprising a test rig capable of holding 12 smartphones for parallel testing, a method for generating synthetic PPG test videos with controllable HR and signal quality, and a host machine for coordinating video playback and data logging. The system achieved a mean absolute percentage error (MAPE) of 0.11\% ± 0.001\% between input and measured HR, and a correlation coefficient of 0.92 ± 0.008 between input and measured PPG signals using a clinically-validated smartphone-based HR app. Bench-testing results of 20 different smartphone models correctly classified all the devices as meeting the ANSI/CTA accuracy standards for HR monitors (MAPE <10\%) when compared to a prospective clinical study with 80 participants, demonstrating high positive predictive value. This platform offers a scalable solution for pre-deployment testing of smartphone HR apps to improve app performance, ensure device compatibility, and advance the field of mobile health.
\end{abstract}

\begin{document}

\maketitle

\section{Introduction}
The ubiquity of smartphones offers a tremendous opportunity for mobile health sensing technology, but the multiplicity of devices and fragmentation of operating systems make it very challenging for developers to test the accuracy of their health apps across different devices and conditions. In particular, smartphone apps that use the built-in camera for heart rate (HR) assessment are among the most popular health apps, but substantial accuracy differences have been reported between different apps \citep{coppetti2017accuracy}, highlighting the need for better testing. Given that there are over 1,000 different smartphone models in the Android ecosystem alone, it is not practical to manually verify each device serially, or conduct expensive human subject testing for each smartphone model.

Functional testers are available to bench-test the accuracy for medical devices like pulse oximeters \citep{en2011medical} and health wearables (AECG100, Whaleteq, Taiwan) by providing inputs with predictable values of HR and blood oxygen saturation, but no test solutions currently exist for smartphone apps. As such, there is a need for a method that allows rapid bench-testing of smartphone apps to enable the discovery of hardware and software issues prior to deployment.

\begin{figure}[ht]
  \centering
  \includegraphics[width=1.0 \textwidth]{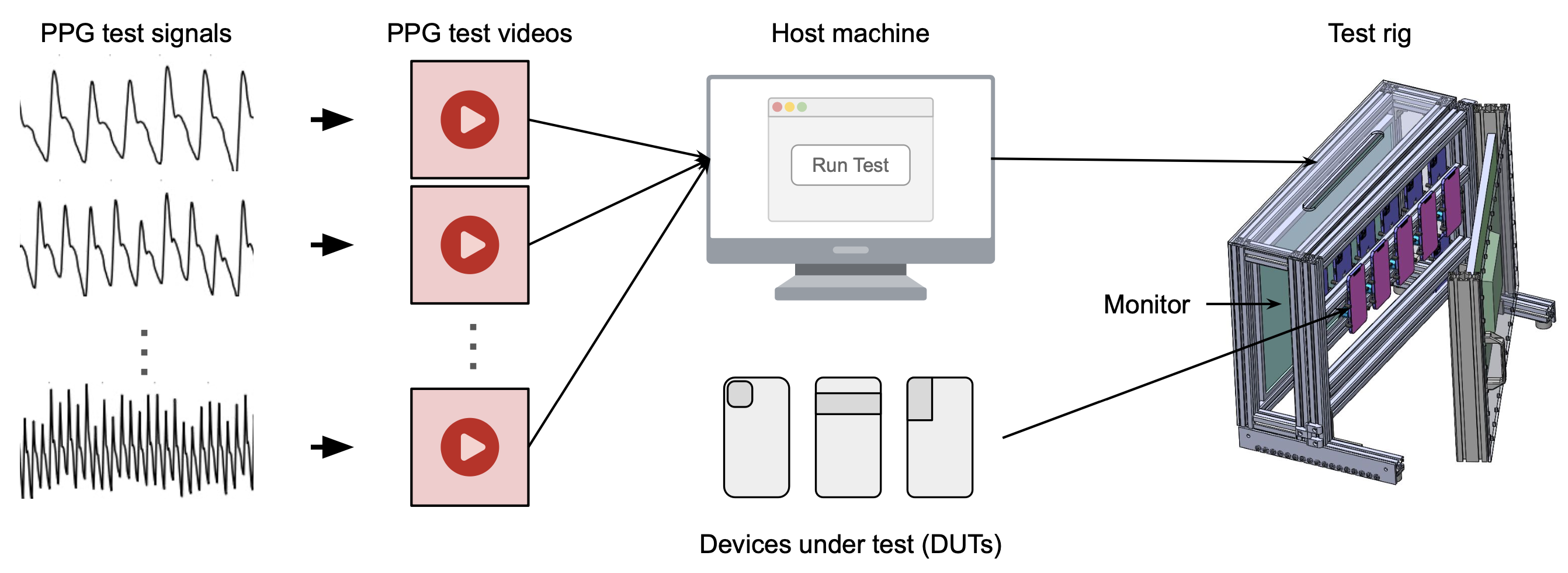}
  \caption{System overview of the platform for bench testing of smartphone apps that use the built-in camera for heart rate measurements using the finger-over-camera technique. \textmd{We created a test set of photoplethysmographic (PPG) waveforms at heart rates ranging from 60 to 180 bpm and generated corresponding synthetic PPG test videos. The smartphones under test are mounted in a test rig with their cameras facing a monitor. A host machine then coordinates the video playback on the monitor and data logging of the devices under test.}}
  \label{fig:system}
\end{figure}

In this work, we present a novel platform that enables high-throughput bench-testing of smartphone apps that use the built-in camera for performing HR measurements via the finger-over-camera technique. These apps measure HR based on video imaging photoplethysmography (iPPG) from the fingertip when placed over the camera by analyzing variations in video frame (skin) color that correspond to changes in blood volume due to the cardiac cycle. First, we designed a test rig that can present multiple smartphones with predictable input PPG signals via video playback. Second, we developed a method to generate synthetic videos encoded with PPG signals to serve as test cases. We validated the platform using a reference smartphone and HR app that was previously clinically validated \citep{bae2022prospective}. Finally, we demonstrate that our system provides a high positive predictive value in predicting the accuracy of different smartphone models in a prospective clinical study.

\section{Overview of System}
\begin{figure}[ht!]
\centering
\includegraphics[width=1.0 \columnwidth]{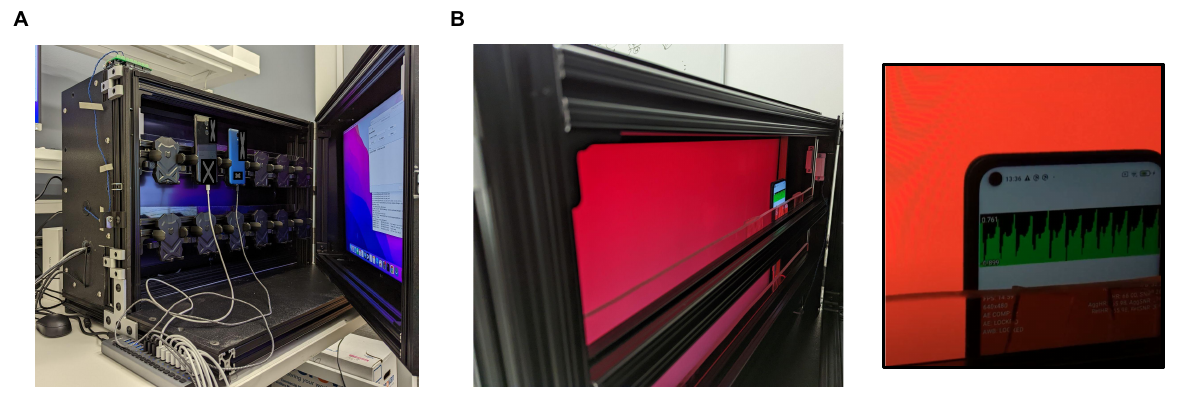}
\caption{Experimental setup for bench testing. \textmd{A. The test rig that comprises an enclosed box with a monitor mounted on the door and horizontal beams with mounts for the smartphones under test. B. Example of a PPG test video being played back to a smartphone. Inset shows a close up of the PPG waveform (black with green background) received by the phone.}}
\label{fig:rig}
\end{figure}

Our overall goal was to design a system that could present a desired and predictable PPG waveform to a smartphone camera, similar to how functional testers work for pulse oximeters and wearable health devices. To this end, our system comprises a method to create test videos from a set of desired PPG waveforms to serve as test inputs, a test rig that houses the devices under test (DUTs) facing a monitor that provides playback of the test videos, and a host machine that coordinates the DUTs and video playback for simultaneous testing (Figure \ref{fig:system}). The test rig is an enclosed box with a monitor mounted on a swinging door and allows up to 12 smartphones to be mounted across horizontal beams inside (Figure \ref{fig:rig}). The swinging door allows easy access to the smartphone rack inside, so technicians can easily swap devices. A hole on the side with a grommet allows cables to pass through to the smartphones and simulator. This design can be expanded to incorporate another monitor on the opposite door to further increase the number of DUTs that can be tested.

\begin{figure*}[h]
	\centering
	\includegraphics[width=1.0 \textwidth]{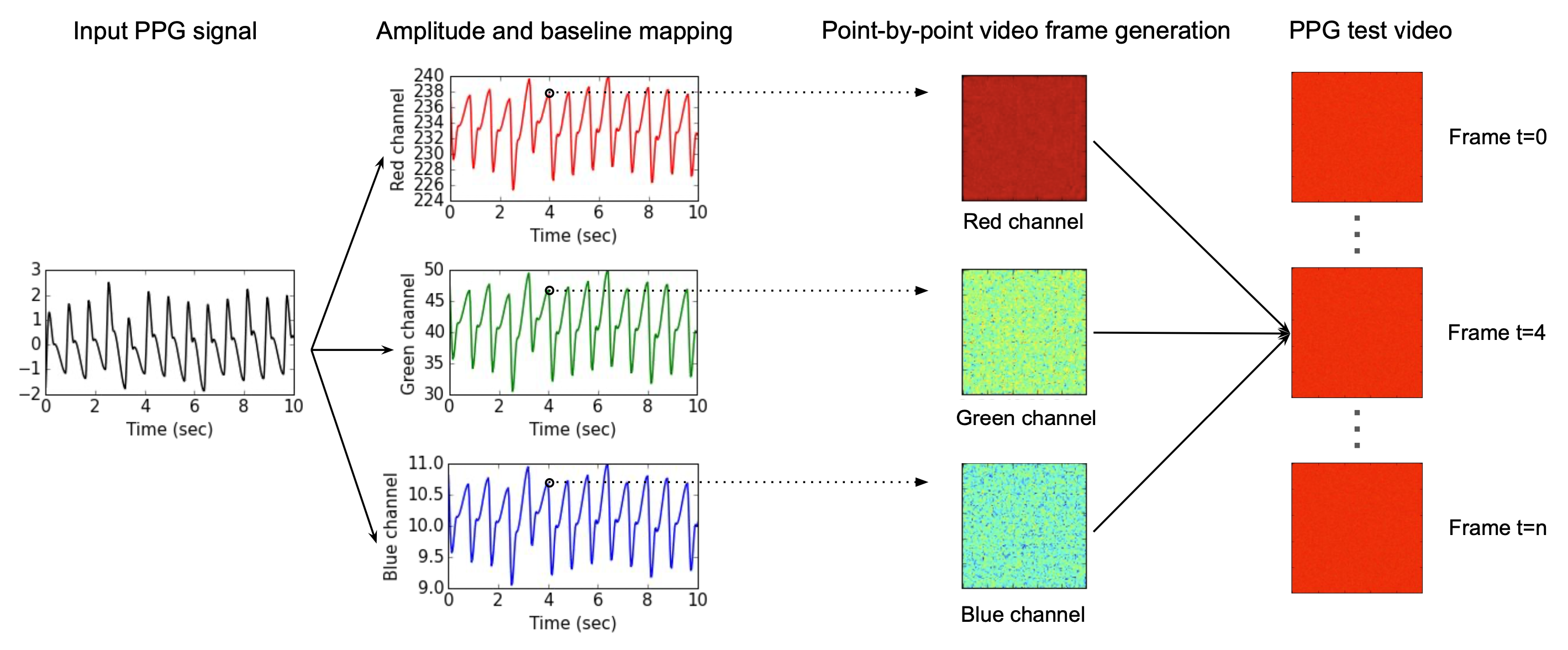}
	\caption{Method to generate synthetic PPG test videos. \textmd{We mapped the target PPG waveform to the appropriate values in the RGB color space and generated a video frame for each data point such that the spatially-averaged mean of the video frames matched the corresponding data point.}}
	\label{fig:ppg_video}
\end{figure*}

We developed a method to create test videos given a PPG waveform that simulated actual videos captured from a finger (Figure \ref{fig:ppg_video}). Here, we used the NeuroKit PPG simulator \citep{makowski2021neurokit2} because it offered control over the desired HR, duration, respiratory sinus arrhythmia (RSA) modulation, baseline drift, motion artifacts, powerline artifacts, etc. In practice, one could also utilize any PPG waveforms, including those available in large public datasets such as the MIMIC-III Waveform database \citep{johnson2016mimic, goldberger2000physiobank}. Given a target PPG waveform, we first mapped the target PPG to the right pixel values in the RGB color space by downsampling to the desired video frame rate, inverting, and rescaling the waveform to the desired range of pixel values, yielding a mapped PPG waveform. This step allowed us to simulate the video color distribution and PPG signal strength under different conditions e.g. skin pigmentation, lighting conditions etc. by using data from actual finger-over-camera videos captured under those conditions from a previous study \citep{bae2022prospective} to inform the correct range of RGB pixel values. We then generated a series of video frames (comprising the 3 color channels), one for each data point in the mapped PPG waveform. Typical pixel values are 8-bit integers (0-255) whereas the PPG signal requires greater precision to preserve information on timing and morphology within a narrow range of pixel values (e.g. 1-2 pixel values in the blue channel). Hence for a given PPG value, we want the spatially-averaged mean of the video frame (comprising pixel values of 8-bit integers) to match the corresponding floating point values in RGB color space computed in the mapped PPG waveform. For each PPG value in RGB color space, we create an $n\times m$ matrix with the desired video frame size using random 8-bit integers drawn from a Gaussian distribution with the mean value set to the desired (floating point) value. We repeated this process for each of the RGB color channels, resulting in three $n\times m$ matrices (one for each color channel) per mapped PPG value. We stacked the three matrices to form a single color frame and the time series of color frames formed the output video containing the desired PPG information.

\section{Validation of System}
\begin{figure}[h]
\centering
\includegraphics[width=1.0 \columnwidth]{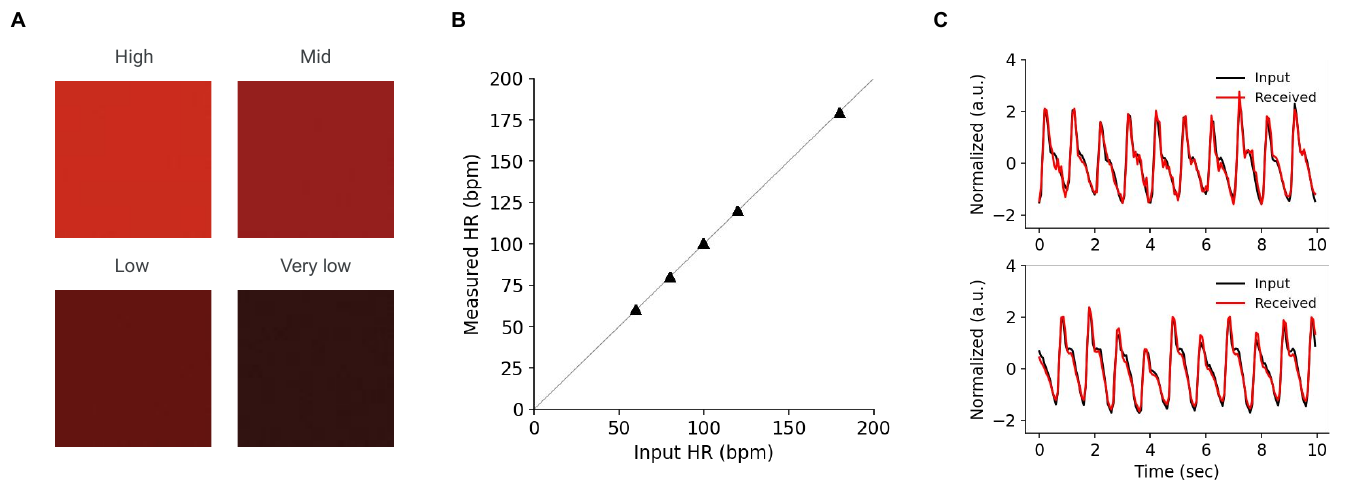}
\caption{A. Example video frames of the synthetic PPG test videos at four different PPG signal strengths representing different levels of brightness and pulse amplitudes. B. Scatter plot of the expected (test input) and measured HR using the reference smartphone. C. Examples of the input PPG waveforms (black line) at very low (top) and high (bottom)signal strengths encoded in the test videos and corresponding PPG waveforms recorded (red line) bya reference smartphone that was previously validated for accurate heart rate (HR) measurements.}
\label{fig:video_ex}
\end{figure}

First, we created a set of 20 synthetic PPG test videos at HRs of 60, 80, 100, 120, 180 bpm, each at four different PPG signal strengths representing different levels of brightness and pulse amplitudes (Figure \ref{fig:video_ex}A). Next, we validated the accuracy of test videos by performing measurements using the clinically-validated HR measurement feature of the Google Fit app \citep{googlefitapp} (study app) running on a reference smartphone, Pixel 3 (Google, Mountain View, CA). The study app was modified to enable triggers from the host computer and automated data logging. The Google Fit HR measurement feature has been validated on the Pixel 3 to measure HR via placing a finger over the back camera with mean absolute percentage errors (MAPEs) of $1.77\% \pm 4.46\%$, $1.32\% \pm 3.30\%$, and $1.77\% \pm 4.77\%$, respectively, for light, medium, and dark skin pigmentation groups \citep{bae2022prospective}. We placed a Pixel 3 on the bench-top rig and performed 20 rounds of testing. In each round, the 20 test videos were played back sequentially. For each test video, we compared the expected (input) HR and the HR the app produced in terms of the absolute percentage error. We also computed the cross correlation between the expected PPG signal embedded in the input test video and the PPG signal recorded by the app. 

The overall MAPE of our system (i.e. HR used to generate the synthetic videos compared to the HR measure by the smartphone app), calculated from 400 paired HR measurements, was $0.11\% \pm 0.001\%$, with a Pearson correlation coefficient of 1.0, indicating that the system simulated different HRs with high accuracy (Figure \ref{fig:video_ex}B). The overall correlation coefficient between input and received PPG waveform signals was 0.92 + 0.008, showing that the system produced simulated PPG signals with accurate rate, rhythm and morphology (Figure \ref{fig:video_ex}C). Over the 20 runs, the coefficients of variation (CoVs) for MAPE and correlation coefficients were 1.17\% and 0.90\%, respectively, indicating high reproducibility. Taken together, these results demonstrate that the system provided accurate and consistent PPG simulations across a wide range of HRs.

\section{Prospective Clinical Validation}
\begin{figure}[t!]
\centering
\includegraphics[width=1.0 \columnwidth]{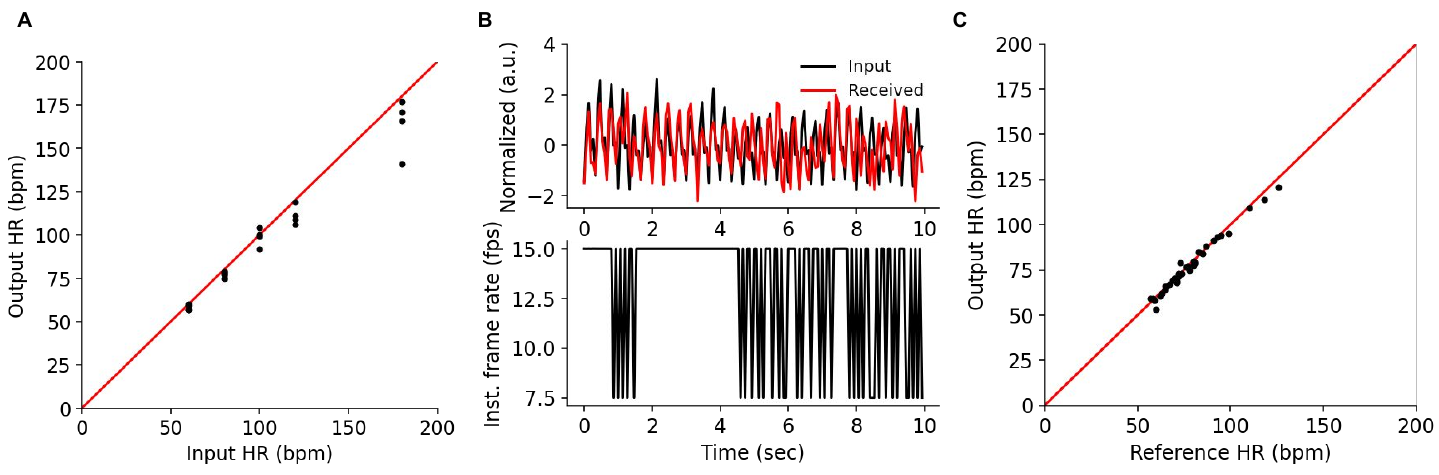}
\caption{Heart rate accuracy on the LG Nexus 5X. \textmd{A. Bench testing performance. Scatterplot of the expected (test input) and measured HR showing larger errors at high heart rate (HRs) of 180 bpm B. Example of the PPG waveform at 180 bpm recorded (red line) by the device and the corresponding instantaneous frame rate showing frequent frame drops that result in a corrupted PPG waveform. Black line indicates the input test waveform. C. Clinical testing performance. Scatterplot of the reference HR and measured HR showing a maximum HR of 126 bpm.}}
\label{fig:nexus-fps}
\end{figure}

After verifying that our system works, we bench-tested the accuracy of 20 different smartphone models representing some of the most popular devices in 2021 from 7 different smartphone manufacturers (Figure \ref{fig:phones}). These smartphones were commercially available devices that did not require any hardware modifications in order to run the study app. We considered a smartphone model accurate if the MAPE was <10\% in accordance with the ANSI/CTA standards for consumer HR monitors \citep{consumer2018physical}, which is based on the ANSI/AAMI standard for HR accuracy for ECG monitors \citep{monitors2002heart}. We first placed the 20 different smartphone models running the Google Fit HR app on the benchtop rig and then played back the 20 test videos sequentially. All smartphone models had a MAPE significantly <10\% (ranging from 0.11 to 5.19\%). We observed that the LG Nexus 5X had the highest MAPE (5.19\%). The errors for the LG Nexus 5X occurred at higher HRs (Figure \ref{fig:nexus-fps}A) and coincided with frequent frame drops (Figure \ref{fig:nexus-fps}B). 

Finally, we evaluated whether our system could accurately classify smartphone models as either accurate or not in measuring HR in actual human subject testing. To this end, we conducted a prospective clinical study to evaluate the accuracy of the same 20 smartphone models. The study protocol was approved by an Institutional Review Board (Advarra, Columbia MD). We obtained informed consent from all participants, and the study was conducted in accordance with the principles of the Declaration of Helsinki. Each participant completed independent 40 HR measurements in total, 36 at rest and 4 post-exercise, using nine different smartphone models randomly selected out of the 20, in such a manner that 40 paired measurements were obtained with each model. During the exercise session, participants used a stationary desk cycle to exercise to 75\% of their maximal HR as tolerated against light to medium resistance, as chosen by the participant. We used an FDA-cleared finger pulse oximeter (Onyx II 9560, Nonin, Plymouth, MN) to provide reference HR measurements. 

\begin{table}[b]
\centering
\caption{Demographics of study participants}
\begin{tabular}{lrr}
\midrule
No. participants &   & 74 \\
Females (\%) &   & 19 (25.7\%) \\
Age, years (mean $\pm$ STD) &   & 47.3 $\pm$ 16.3 \\
Age group & $<$40 years & 29 (39.2\%) \\
  & 40-59 years & 20 (27.0\%) \\
  & $>$59 years & 25 (33.8\%) \\
Fitzpatrick skin type & Type I-III & 27 (36.5\%) \\
  & Type IV-V & 24 (32.4\%) \\
  & Type VI  & 23 (31.1\%) \\
Heart rate, bpm (mean $\pm$ STD) &   & 76.9 $\pm$ 13.7 \\  
\bottomrule
\end{tabular}
\label{table:participants}
\end{table}

\begin{figure}[h]
\centering
\includegraphics[width=1.0 \columnwidth]{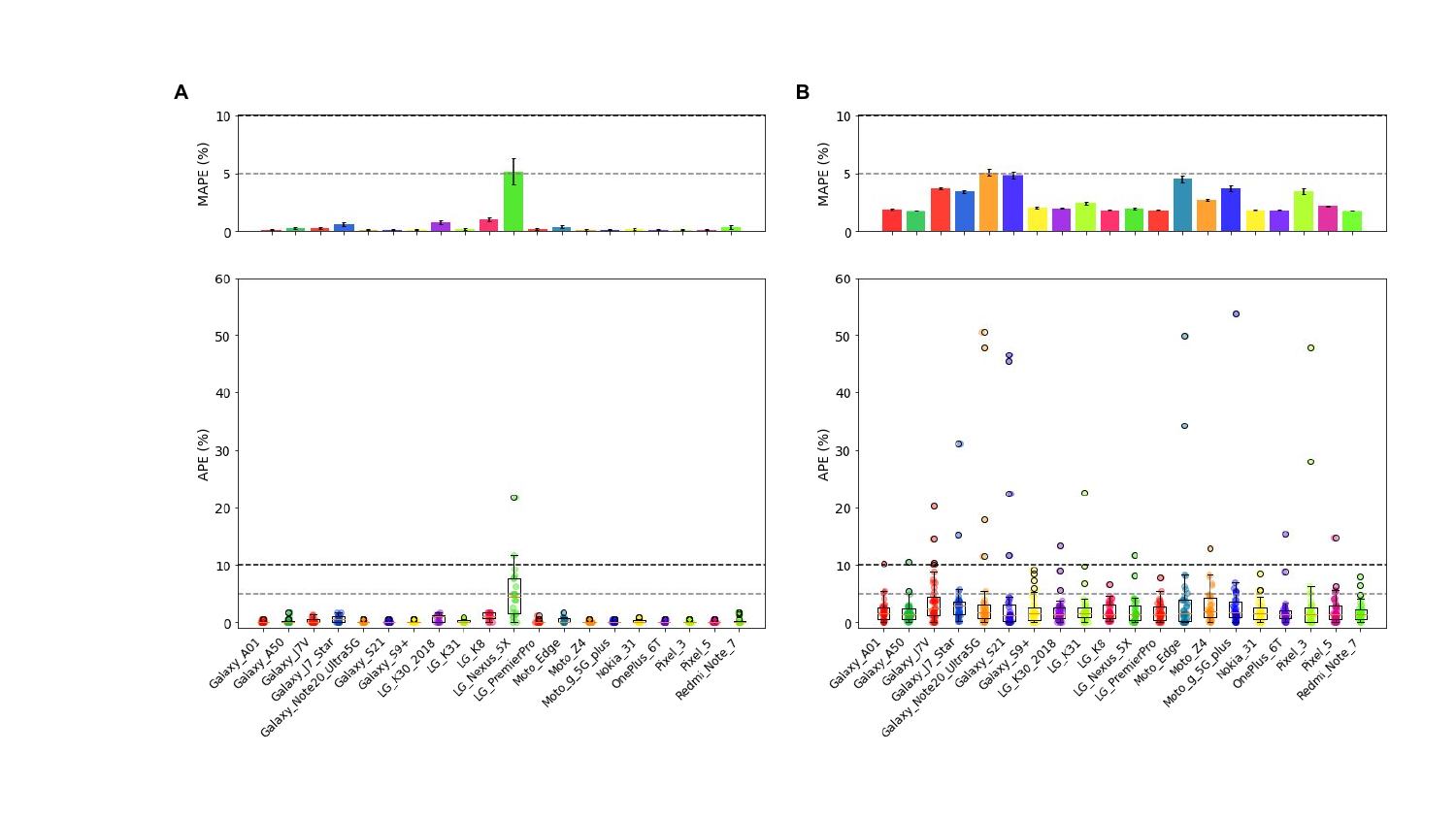}
\caption{Evaluating the accuracy of various smartphone models. \textmd{(A) Bench-testing results of heart rate (HR) measurement accuracy in terms of mean absolute percentage errors (MAPEs) and corresponding boxplots of the absolute percentage errors (APEs) across 20 different devices. (B) HR measurement accuracy of the same devices from a prospective clinical study with consented participants.}}
\label{fig:clinical_accuracy}
\end{figure}

A total of 74 participants were included in our analysis (six participants were excluded because they did not have reference HR data due to failure of the pulse oximeter). These participants represented a wide range of ages and skin tones (Table \ref{table:participants}). HR measurements ranged from 53 to 147 bpm. In general, the MAPEs in the clinical study (1.74-5.11\%) were slightly higher than those observed during bench-testing across smartphone models (Figure \ref{fig:clinical_accuracy}). There were also more high-error outliers (Figure \ref{fig:clinical_accuracy}B). This might be because the phones were handheld in this study, which introduced motion artifacts. Nonetheless, all 20 smartphone models produced a MAPE significantly <10\% against the pulse oximeter, meeting the ANSI/CTA accuracy standards for HR monitors. This indicated that our system correctly classified all the smartphone models as accurate. 

\begin{figure}[h]
\centering
\includegraphics[width=0.6 \columnwidth]{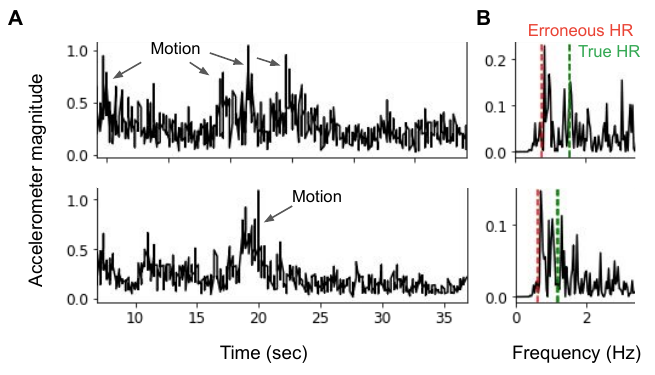}
\caption{Phone motion during Galaxy Note 20 Ultra measurements. \textmd{(A) Time series of the accelerometer magnitude during the two measurements with large heart rate (HR) errors. Arrows indicate phone movement. (B) The dominant frequency in the accelerometer traces corresponded to the frequency of the erroneous HR predictions (red dotted lines). Green dotted lines indicate the frequency of the true HR.}}
\label{fig:accel-motion}
\end{figure}

Surprisingly, the LG Nexus 5X had a lower MAPE (1.96\%) in the clinical study than observed during bench-testing (5.19\%). This might be due in part to the lower range of HRs measured in the clinical study, as the errors observed during bench-testing occurred at higher HRs, particularly those at 180 bpm (Figure \ref{fig:nexus-fps}A). In contrast, the highest HRs recorded in the clinical study involving this phone were around 125 bpm (Figure \ref{fig:nexus-fps}C). The Galaxy Note 20 Ultra had the highest MAPE (5.11\%) due to two large outliers (Figure \ref{fig:clinical_accuracy}B). We examined the phone accelerometer traces and observed that there was phone motion during those two measurements, suggesting that motion artifacts may be the cause of the erroneous readings (Figure \ref{fig:accel-motion}A). Indeed, the peaks in the frequency spectrum of the accelerometer tracings coincided with the frequency of the erroneous HRs (Figure \ref{fig:accel-motion}B).

\section{Discussion}
To our knowledge, this is the first demonstration of a bench-testing system that provides high-throughput verification of smartphone apps that use built-in cameras for HR measurement. Our system was able to playback synthetic PPG test videos at different HRs with accurate rate, rhythm and morphology. In its current design, up to 12 phones could be tested simultaneously.

When relying on manual testing using humans, it is typically difficult to capture infrequent HR values, such as extreme bradycardia or tachycardia. One advantage of our system is the ability to synthesize PPG waveforms of any target HR, which enables testing across the full range of physiologically plausible HRs. The method we developed to create synthetic PPG test videos is also agnostic to the origin of the PPG recordings; both simulated and real PPG waveforms can be used as the desired input source. This flexibility allows incorporating real PPG waveforms exhibiting abnormal heart rhythms such as atrial fibrillation to be used for creating test videos.

Importantly, our system allows repeatable testing using a standardized set of test videos. This provides app developers with a method to quickly test, iterate, and improve their algorithms for HR estimation. For example, developers can use bench-testing to measure performance regression for each release version of their HR algo models, or evaluate compatibility with a new smartphone model. In contrast, human-based manual testing does not allow controlling for the same input PPG waveform as HR varies naturally from beat to beat.

An alternative approach is for developers to evaluate their smartphone HR algorithms offline using a reference database of PPG waveforms. However, this approach would only be able to evaluate correct implementation of the algorithms, but would not test the signal acquisition pathway via smartphone hardware-specific components such as the camera and the image signal processor (ISP). Potential issues that our system might be able to identify beyond bugs in the on-device algorithm include ISP misconfigurations, unstable video frame rates, poor camera signal quality and the like.

Our work has some limitations. First, our system is not able to simulate the mechanical-optical interface between the smartphone camera and finger. As such, it is not able to identify potential issues related to human factors such as smartphone ergonomics that may increase motion artifacts if a particular hardware design makes it difficult for an individual to hold the phone steady throughout the measurement duration, or noisy measurements due excessive finger pressure against the camera causing vasoconstriction. In our study, we only bench-tested the smartphones using clean PPG waveforms, hence the MAPEs observed in the clinical study with human participants were higher. Nonetheless, developers could test the robustness of their HR algorithms by using real PPG waveforms containing motion artifacts to generate test videos. One could also inject synthetic noise at a desired signal-to-noise ratio into the PPG waveforms used to generate the test videos. In addition, our system is not able to identify potential issues related to the smartphone flashlight such as suboptimal LED wavelength, or LED noise, if it is required to provide additional illumination as is employed by some finger-based heart rate measurement apps. Lastly, as none of the 20 different phones were found to be inaccurate for HR measurements in our prospective clinical study, we were unable to evaluate the negative predictive value of our platform. Future work could involve bench-testing an even larger sample of smartphones to identify such devices for subsequent clinical evaluation.

\section{Conclusion}
We presented the design, implementation and evaluation of a novel, high-throughput platform for bench-testing smartphone-based HR measurement apps, directly addressing the challenges of device fragmentation and the lack of standardized testing in the mobile health domain. Our system's ability to simultaneously evaluate multiple devices, coupled with its precise control over simulated physiological signals, provides a significant advancement over existing methods. This work enables developers to rigorously assess app performance and device compatibility before deployment, leading to more robust and reliable mobile health solutions. 

\bibliography{references}
\end{document}